\documentclass{article} 
\usepackage{iclr2025_conference,times}
\usepackage{graphicx}
\usepackage{booktabs}
\usepackage{multirow}

\usepackage{amsmath,amsfonts,bm}

\def\eqref#1{equation~\ref{#1}}

\def\1{\bm{1}}

\DeclareMathAlphabet{\mathsfit}{\encodingdefault}{\sfdefault}{m}{sl}
\SetMathAlphabet{\mathsfit}{bold}{\encodingdefault}{\sfdefault}{bx}{n}

\usepackage{hyperref}
\usepackage{url}
\usepackage{graphicx}
\usepackage{amsmath}
\usepackage{amssymb}
\usepackage{booktabs}
\usepackage{relsize}
\usepackage{float}
\usepackage{adjustbox}
\usepackage[bottom]{footmisc}

\usepackage[capitalize]{cleveref}
\crefname{section}{Sec.}{Secs.}
\Crefname{section}{Section}{Sections}
\Crefname{table}{Table}{Tables}
\crefname{table}{Tab.}{Tabs.}

\newcommand{\wala}{WaLa}
\newcommand\blfootnote[1]{%
  \begin{NoHyper}%
  \renewcommand\thefootnote{}\footnote{#1}%
  \addtocounter{footnote}{-1}%
  \end{NoHyper}%
}

\title{Wavelet Latent Diffusion (\wala): Billion-Parameter 3D Generative Model with Compact Wavelet Encodings}

\author{
Aditya Sanghi \And  Aliasghar Khani$^{*}$ \And  Pradyumna Reddy$^{*}$ \And  Arianna Rampini  
\AND
Derek Cheung \And Kamal Rahimi Malekshan \And    Kanika Madan \And   Hooman Shayani
\AND
\centerline{\color{magenta} \href{https://autodeskailab.github.io/WaLaProject}{https://autodeskailab.github.io/WaLaProject}}
}

\iclrfinalcopy 
\begin{document}


\maketitle
\blfootnote{* Equal contribution. For further inquiries, please email \href{mailto:aditya.sanghi@autodesk.com}{aditya.sanghi@autodesk.com} }
        

\vspace{-0.99cm}
\begin{abstract}
Large-scale 3D generative models require substantial computational resources yet often fall short in capturing fine details and complex geometries at high resolutions. We attribute this limitation to the inefficiency of current representations, which lack the compactness required to model the generative models effectively. To address this, we introduce a novel approach called \textbf{Wa}velet \textbf{La}tent Diffusion, or \textbf{WaLa}, that encodes 3D shapes into a wavelet-based, compact latent encodings. Specifically, we compress a $256^3$ signed distance field into a $12^3 \times 4$ latent grid, achieving an impressive 2,427× compression ratio with minimal loss of detail. This high level of compression allows our method to efficiently train large-scale generative networks without increasing the inference time. Our models, both conditional and unconditional, contain approximately one billion parameters and successfully generate high-quality 3D shapes at $256^3$ resolution. Moreover, WaLa offers rapid inference, producing shapes within two to four seconds depending on the condition, despite the model’s scale. We demonstrate state-of-the-art performance across multiple datasets, with significant improvements in generation quality, diversity, and computational efficiency. We open-source our code and, to the best of our knowledge, release the largest pretrained 3D generative models across different modalities: \href{https://github.com/AutodeskAILab/WaLa}{\textbf{\textcolor{magenta}{https://github.com/AutodeskAILab/WaLa}}}.

\begin{figure*}[h]
    \begin{center}
        \includegraphics[width=\linewidth]{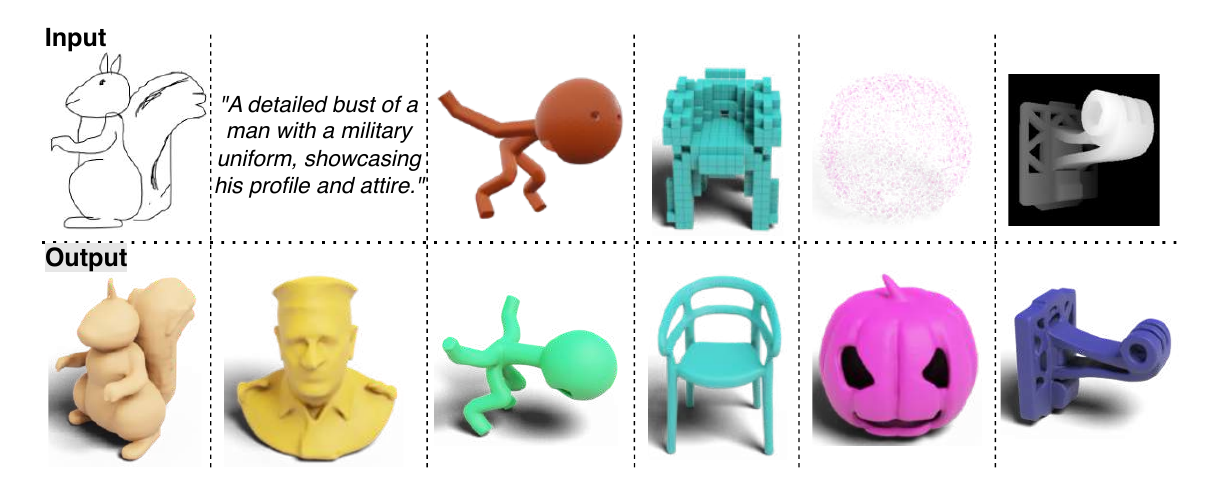}
         
        \caption{We propose a new 3D generative model, called \wala, that can generate shapes from conditions such as sketches, text, single-view images, low-resolution voxels, point clouds \& depth-maps.}
        \label{fig:teaser}
    \end{center}
\end{figure*}

\end{abstract}


\vspace{-0.99cm}

\section{Introduction}
\vspace{-0.2cm}
Training generative models on large-scale 3D data presents significant challenges. The cubic nature of 3D data drastically increases the number of input variables the model must handle, far exceeding the complexity found in image and natural language tasks. This complexity is further compounded by storage and streaming issues. Training such large models often requires cloud services, which makes the process expensive for high-resolution 3D datasets as these datasets take up considerable space and are slow to stream during training. Additionally, unlike other data types, 3D shapes can be represented in various ways, such as voxels, point clouds, meshes, and implicit functions. Each representation presents different trade-offs between quality and compactness. Determining which representation best balances high fidelity with compactness for efficient training and generation remains an open challenge. Finally, 3D representations often exhibit complex hierarchical structures with details at multiple scales, making it challenging for a generative model to capture both global structure and fine-grained details simultaneously.

\begin{figure*}[!t]
    \centering
    \includegraphics[width=0.87\linewidth]{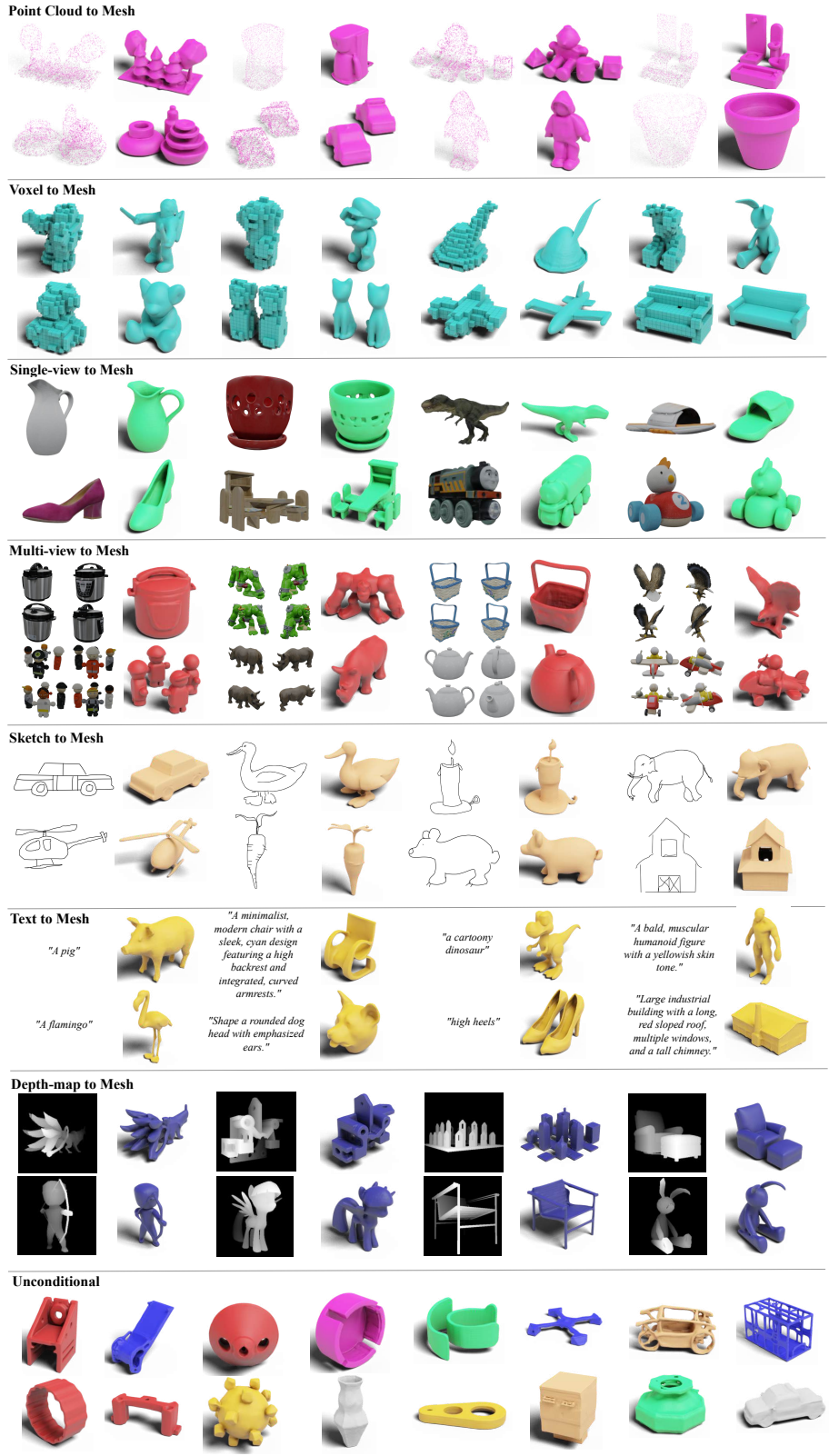}
    \vspace{-0.2cm}
    \caption{WaLa generates 3D shapes across various input modalities (see appendix for more).}
    \label{fig:teaser_2}
    \vspace{-1pt}
\end{figure*}

To address these challenges, current state-of-the-art methods for large generative models typically employ three main strategies. The first strategy involves using low-resolution representations, such as sparse point clouds ~\citep{point-e, shape-e}, low-polygon meshes ~\citep{chen2024meshanything}, or coarse grids ~\citep{cheng2023sdfusion, sanghi2023sketch}. While these approaches reduce computational complexity, they are limited in their ability to model the full distribution of 3D shapes, struggle to capture intricate details, and often lead to lossy representations. The second approach represents 3D shapes through a collection of 2D images ~\citep{yan2024object} or incorporates images ~\citep{hong2023lrm, li2023instant3d, liu2024one, xu2023dmv3d, siddiqui2024meta, bensadoun2024meta} into the training loss. However, this method suffers from long training times due to the need for rendering and can fail to capture internal details of 3D shapes, as it primarily focuses on external appearances. The third strategy introduces more compactness  into the input representations ~\citep{hui2024make, zhou2024udiff, ren2024xcube, yariv2024mosaic, xiong2024octfusion, zhang2024clay} to reduce the number of variables the generative model must handle. While these representations can be sparse ~\citep{ren2024xcube, yariv2024mosaic, xiong2024octfusion}, they are often irregular or discrete in nature making it challenging to be modeled via neural networks and can still be relatively large compared  to image or natural language data ~\citep{hui2024make, zhou2024udiff}, thus making it difficult to scale the model parameters efficiently.

One prominent compact input representation is wavelet-based representation, which includes Neural Wavelet~\citep{hui2022neural}, UDiFF~\citep{zhou2024udiff}, and wavelet-tree frameworks~\citep{hui2024make}. These methods utilize wavelet transforms and their inverses to seamlessly convert between wavelet spaces and high-resolution truncated signed distance function (TSDF) representations. They offer several key advantages: data can be easily compressed by discarding selected coefficients with minimal loss of detail, and the interrelationships between coefficients facilitate efficient storage, streaming, and processing of large-scale 3D datasets compared to directly using TSDFs~\citep{hui2024make}. However, despite these benefits, wavelet-based representations remain substantially large, especially when scaling up for large-scale generative models. For example, a $256^3$ TSDF can be represented as a wavelet-tree of size $46^3 \times 64$~\citep{hui2024make}, which is equivalent to a $1440 \times 1440$ RGB image. Scaling within this space continues to pose significant challenges.

In this work, we build upon the wavelet representation described above and introduce the \emph{\textbf{Wa}velet \textbf{La}tent Diffusion} (\textbf{\wala}) framework. This framework further compresses the wavelet representation to obtain compact latent encodings without significant information loss, thereby efficiently enabling us to scale a diffusion-based generative model within this space.  Starting with a truncated signed distance function (TSDF) of a shape, we first convert it into 3D wavelet tree representation as in ~\citet{hui2024make}. Then, we train a convolution-based VQ-VAE model with adaptive sampling loss and \textit{balanced fine-tuning} to compress a $256^3$ TSDF into a $12^3 \times 4$ grid, achieving a remarkable $2,427\times$ compression ratio while maintaining an impressive reconstruction without a significant loss of detail. For example, as shown in Table~\ref{tab:rep_compare}, an Intersection over Union (IOU) of 0.978 is achieved on the GSO dataset. Compared to other representations, this approach requires fewer input variables for the generative model while retaining high reconstruction accuracy. Consequently, the generative model does not need to model local details and can focus on capturing the global structure. Moreover, by significantly reducing the number of input variables that the generative model must handle due to this compression, we enable the training of large-scale 3D generative models with up to a billion parameters, producing highly detailed and diverse shapes. \wala ~also supports controlled generation through multiple input modalities without adding significant inductive biases, making the framework flexible and adaptable beyond single-view 3D reconstruction tasks. As a result, our model generates 3D shapes with complex geometry, plausible structures, intricate topologies, and smooth surfaces. This is demonstrated in Figures~\ref{fig:teaser}
 and~\ref{fig:teaser_2}, where high-quality 3D meshes can be obtained by applying marching cubes to the SDF generated from different input modalities such as text, sketch, low-resolution voxel, point cloud, single-view, and multi-view images.



In summary, we make the following contributions:

\begin{itemize}
    \item We introduce a \emph{\textbf{Wa}velet \textbf{La}tent Diffusion} (\textbf{\wala}) framework that tackles the dimensional and computational challenges of 3D generation with impressive compression while maximizing fidelity.

    \item Our large billion-parameter model generates high-quality 3D shapes within two to four seconds, significantly outperforming state-of-the-art benchmarks in 3D shape generation.
    
    \item Our model demonstrates exceptional versatility, accepting diverse input modalities such as single/multi-view images, voxels, point clouds, depth data, sketches, and textual descriptions (see Figure ~\ref{fig:teaser} and ~\ref{fig:teaser_2}), making it applicable to a wide range of 3D modeling tasks.
    
    \item To foster reproducibility and stimulate further research in this domain, we release what we believe is, to the best of our knowledge, the largest 3D generative model to date that works across various input modalities, comprising approximately one billion parameters. The model is available at \href{https://github.com/AutodeskAILab/WaLa}{\textbf{https://github.com/AutodeskAILab/WaLa}}.
\end{itemize}

\begin{table}[h]
\caption{
3D representations compared on GSO dataset~\citep{downs2022google}: Intersection over Union (IoU) for accuracy \& number of input variables for generative models to evaluate complexity.
}
\label{tab:rep_compare}
\centering
{\small
\resizebox{0.65\linewidth}{!}{
\begin{tabular}{c|c|c}
\toprule
Representation            & IoU & Number of Input Variables \\
\midrule
Ground-truth SDF ($256^3$) & $1.0 $ & $16,777,216$ ($\sim64\text{MB}$)         \\ 
Point Cloud~\citep{nichol2022point}  & $0.8642$ & $12,288$ ($\sim0.05\text{MB}$)   \\
Latent Vectors~\citep{jun2023shap} & $0.8576$ & $1,048,576$ ($\sim4\text{MB}$)   \\
Coarse Component~\citep{hui2022neural}  & $0.9531$  & $97,336$ ($\sim0.4\text{MB}$)           \\

Wavelet tree~\citep{hui2024make}   & $0.9956$ & $1,129,528$ ($\sim4.3\text{MB}$)   \\ 
\midrule
\textbf{\wala} & $0.9780 $  & $6,912$ ($\sim0.03\text{MB}$)         \\ 
\bottomrule
\end{tabular}}}
\end{table}

\vspace{-0.2cm}
\section{Related Work}
\vspace{-0.2cm}
\noindent \textbf{Neural Shape Representations.}  
Several representations have been explored for Deep learning for 3D data. Initially, volumetric methods using 3D convolutional networks were employed~\citep{wu20153d, maturana2015voxnet}, but they were limited by resolution and efficiency. The field then advanced to multi-view CNNs that apply 2D processing to rendered views~\citep{su2015multi, qi2016volumetric}, and further explored sparse point cloud representations with networks like PointNet and its successors~\citep{qi2017pointnet, qi2017pointnet++, wang2019dynamic}. Additionally, neural implicit representations for compact, continuous modeling were developed~\citep{park2019deepsdf, mescheder2019occupancy, chen2019learning}. Explicit mesh-based and boundary representations (BREP) have gained attention, enhancing both discriminative and generative capabilities in CAD-related applications~\citep{hanocka2019meshcnn, chen2024meshanything, jayaraman2021uv, lambourne2021brepnet}. Recently,  wavelet representations ~\citep{hui2022neural, zhou2024udiff, hui2024make} have become popular. 
Wavelet decompositions of SDF signals enable tractable modeling of high-resolution shapes.
In this work, we extend the previous research by addressing the dimensional and computational hurdles of 3D generation. Our novel techniques for efficient shape processing enable high-quality 3D generation at scale, accommodating datasets with millions of shapes.

\noindent \textbf{3D Generative Models.} 
3D generative models have evolved rapidly, initially dominated by Generative Adversarial Networks (GANs) \citep{goodfellow2014generative, wu2016learning}. Subsequent advancements integrated differentiable rendering with GANs, utilizing multi-view losses for enhanced fidelity \citep{chan2022efficient}. Parallel developments explored normalizing flows~\citep{yang2019pointflow, klokov2020discrete, sanghi2022clip} and Variational Autoencoders (VAEs)~\citep{mo2019structurenet}. Additionally, autoregressive models also gained traction for their sequential generation capabilities~\citep{cheng2022autoregressive, nash2020polygen, sun2020pointgrow, mittal2022autosdf, yan2022shapeformer, zhang20223dilg, sanghi2023clip}.
The recent success of diffusion models in image generation has sparked a great interest in their application to 3D contexts. Most current approaches employ a two-stage process: first,  a Vector-Quantized VAE (VQ-VAE) on 3D representations such as triplanes~\citep{shue20233d, chou2023diffusion, peng2020convolutional, reddy2024g3dr, siddiqui2024meta, chen2022tensorf,gao2022get3d,NFD}, implicit forms~\citep{zhang20233dshape2vecset, li2023diffusion, cheng2023sdfusion}, or point clouds~\citep{jun2023shap, zeng2022lion} is trained, and then, diffusion models are applied to the resulting latent space. Incorporating autoencoders to process latent spaces allow for the generation of complex representations like point clouds~\citep{jun2023shap, zeng2022lion} and implicit forms~\citep{zhang20233dshape2vecset, li2023diffusion, cheng2023sdfusion, zhang2024clay}. Direct training of diffusion models on 3D representations, though less explored, has shown promise for point clouds~\citep{nichol2022point, zhou20213d, luo2021diffusion, nakayama2023difffacto}, voxels~\citep{zheng2023locally},  occupancy~\citep{ren2024xcube}, and neural wavelet coefficients~\citep{hui2022neural, liu2023exim, hui2024make}.
Our work advances this frontier by bridging the gap between compact representation and high-fidelity generation.

\noindent \textbf{Conditional 3D Models.} 
Two primary paradigms dominate conditional 3D generative models, each with its own approach to 3D content creation.
The first paradigm ingeniously repurposes large-scale 2D conditional image generators, such as ~\citep{rombach2022high} or Imagen~\citep{saharia2022photorealistic}, for 3D synthesis. This approach employs a differentiable renderer to project 3D shapes into 2D images, enabling comparison with target images or alignment with text-to-image model distributions ~\citep{jain2022zero, michel2022text2mesh, poole2022dreamfusion}. Initially focused on text-to-3D generation, this method has expanded to accommodate various input modalities, including single and multi-view images~\citep{deng2023nerdi, melas2023realfusion, xu2022neurallift, liu2023zero, deitke2023objaverse, qian2023magic123, shi2023mvdream,wang2023prolificdreamer,liu2023zero1to3}, and even sketches~\citep{mikaeili2023sked}. 
This approach, while novel, is limited by its computational demands. An alternative paradigm uses dedicated conditional 3D generative models trained on either paired datasets or through zero-shot learning. These paired models show adaptability to various input conditions,
ranging from point clouds~\citep{zhang20223dilg, zhang2023vec} and images~\citep{zhang20223dilg, nichol2022point, jun2023shap, zhang2023vec, LaRa, tang2024lgm, li2023instant3d, xu2024instantmesh, zhang2024clay, siddiqui2024meta, bensadoun2024meta} to low-resolution voxels~\citep{chen2021decor, chen2023shaddr}, sketches~\citep{lun20173d, guillard2021sketch2mesh, gao2022sketchsampler, kong2022diffusion}, and textual descriptions~\citep{nichol2022point, jun2023shap, ren2024xcube, yariv2024mosaic}. Concurrently, zero-shot methods have gained traction, particularly in text-to-3D~\citep{sanghi2022clip, sanghi2023clip, liu2022iss, xu2023dream3d, yan2024omages64} and sketch-to-3D applications~\citep{sanghi2023sketch}, showcasing the potential for more flexible and generalizable 3D generation. 
We expand on the second paired paradigm, developing a large-scale paired conditional generative model for 3D shapes. This approach enables fast generation without per-instance optimization, supports diverse inputs, and facilitates unconditional generation and zero-shot tasks like shape completion.
\vspace{-0.2cm}
\section{Method}
\vspace{-0.2cm}

\begin{figure*}[!t]
    \centering
    \includegraphics[width=\textwidth]{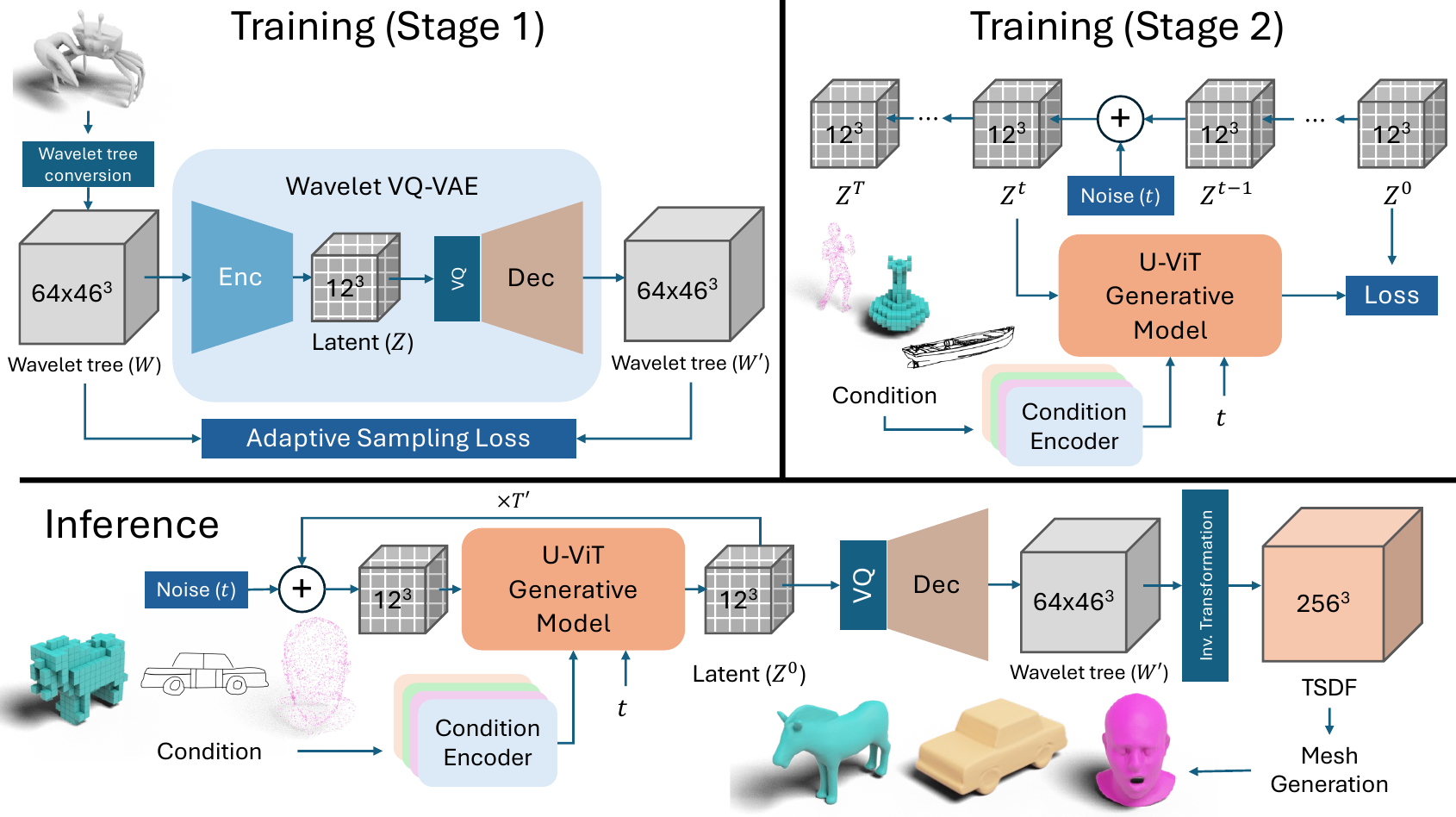}
    \caption{Overview of the \wala~network architecture and 2-stage training process and inference method. Top Left: Stage 1 autoencoder training, compressing diffusible wavelet tree ($W$) shape representation into a compact latent space.
    Top Right: Conditional/unconditional diffusion training.
    Bottom: Inference pipeline, illustrating sampling from the trained diffusion model and decoding the sampled latent into a Wavelet Tree ($W$), then into a mesh.
    }
    \label{fig:architecture}
\end{figure*}

Training generative models on large-scale 3D data is challenging because of the data's complexity and size. This has driven the creation of compact representations like neural wavelets, facilitating efficient neural network training. To represent a 3D shape with wavelets, it is first converted into a Truncated Signed Distance Function (TSDF) grid.
 A wavelet transform is then applied to decompose this TSDF grid into coarse coefficients ($C_0$) and detail coefficients at various levels ($D_0$, $D_1$, $D_2$). Various wavelet transforms, such as Haar, biorthogonal, or Meyer wavelets, can be employed. Most current methods utilize the biorthogonal wavelet transform \citep{hui2022neural, zhou2024udiff, hui2024make}. The coarse coefficients primarily capture the essential shape information, while the detail coefficients represent high-frequency details. To compress this representation, different filtering schemes can be applied to remove certain coefficients, though this involves a trade-off in reconstruction quality. In the neural wavelet representation~\citep{hui2022neural}, all detail coefficients are discarded during the training of the generative model and a regression network is used to predict the missing detail coefficients $D_0$. In contrast, the wavelet-tree representation~\citep{hui2024make} retains all coarse coefficients ($C_0$), discards the third level of detail coefficients ($D_2$), and selectively keeps the most significant coefficients from $D_0$ along with their corresponding details in $D_1$ using a subband coefficient filtering scheme. The neural wavelet representation, while modeling a smaller number of input variables, has lower reconstruction quality than the wavelet-tree representation, making latter a more attractive option.

Building upon these efficient wavelet representations, our method requires a large collection of 3D shapes. Let $\mathcal{S} = \left\{ \left( W_n, \Theta_n \right) \right\}_{n=1}^N$, denote a dataset of $N$ 3D shapes such that each shape $S_n \in \mathcal{S}$ is represented by a diffusible wavelet tree representation $W_n$~\citep{hui2024make} and an optional associated condition $\Theta_n$. The representation $W_n \in \mathbb{R}^{46^3 \times 64}$ is obtained by converting a TSDF of resolution $256^3$. Depending on the conditional generative model, the condition $\Theta_n$ can be a single-view image, multi-view images, a voxel representation, a point cloud, or multi-view depth maps, and may be omitted if the model is unconditional or when training the vector-quantized autoencoder (VQ-VAE). Training our model comprises two stages: First, we train a convolution-based VQ-VAE to encode the diffusible wavelet tree representation into a more compact grid latent space $Z$ using the adaptive sampling loss. At this stage, to further enhance reconstruction accuracy, we fine-tune the VQ-VAE using a simple approach we call \textit{balanced fine-tuning}. This VQ-VAE encodes a latent grid $Z_n \in \mathbb{R}^{12^3 \times 4}$ for each shape $S_n \in \mathcal{S}$. In the second stage, we train a diffusion-based generative model on this latent grid $Z_n$ that can be conditioned on a sequence of condition vectors derived from one of the aforementioned conditions. During inference, we initiate with a completely noisy latent vector and employ the conditional generative network to denoise it progressively through the inverse diffusion process, utilizing classifier-free guidance. The two-step process is detailed in Figure ~\ref{fig:architecture} and is also explained next.

\subsection{Stage 1: Wavelet VQ-VAE}
\label{sec:stage_1_wavelet_vqvae}

Our primary objective is to compress the diffusible wavelet tree representation \citep{hui2024make} into a compact latent space without significant loss of fidelity, thereby facilitating the training of a generative model directly on this latent space. Decoupling compression from generation allows for efficient scaling of a large generative model within the latent space. To this end, we employ a convolution-based VQ-VAE, known for producing sharper reconstructions and mitigating issues like posterior collapse \citep{van2017neural, razavi2019generating, baykal2024edvae}. Specifically, the encoder \( Enc(\cdot) \) maps the input \( W_n \) to a latent representation \( Z_n = Enc(W_n) \), which is then quantized as $\mathrm{VQ}(Z_n)$ via a vector quantization layer and decoded by \( Dec(\cdot) \) to reconstruct the shape \( W_n' = Dec(\mathrm{VQ}(Z_n)) \). By integrating the vector quantization layer with the decoder, as in \citep{rombach2022highresolution}, we ensure that the generative model is trained on pre-quantized latent codes. This approach leverages the robustness of the quantization layer to small perturbations by mapping generated codes to the nearest embeddings in a codebook after generation. Empirical results confirm the effectiveness of this strategy, see Ablation Section \ref{pre_or_quant}.

To train the VQ-VAE, we employ a combination of three losses: a reconstruction loss to ensure fidelity between the original and reconstructed shapes, a codebook loss to encourage the codebook embeddings to adapt to the distribution of encoder outputs, and a commitment loss to align the encoder's outputs closely with the codebook embeddings. 
We apply a reconstruction loss \( \mathcal{L}_{\text{rec}}(W_n, W_n') \),  during which we adopt a  adaptive sampling loss strategy ~\citep{hui2024make} to focus more effectively on high-magnitude detail coefficients (i.e., \( D_0 \) and \( D_1 \)) while still considering the others. Since most detail coefficients are low in magnitude and contribute minimally to the overall shape quality, this approach identifies the significance of these coefficients in each subband based on their magnitude relative to the largest coefficient, forming a set \( P_0 \) of important coordinates. By structuring the training loss to emphasize these crucial coefficients and incorporating random sampling of less important ones, the model efficiently concentrates on key information without neglecting finer details. This is formalized in the equation below:

\begin{equation}
\mathcal{L}_{\text{rec}} = L_{\text{MSE}}(C_0, C'_0) + \frac{1}{2} \sum_{D \in \{D_0, D_1\}} \left[ L_{\text{MSE}}(D[P_0], D'[P_0]) + L_{\text{MSE}}(R(D[P_0']), R(D'[P_0'])) \right]
\end{equation}

In this context, \( L_{\text{MSE}}(X, Y) \) denotes the mean squared error between \( X \) and \( Y \). The coefficients \( C_0, D_0, D_1 \) extracted from \( W_n \) represent the coarse and detail components, respectively, while their reconstructed counterparts \( C'_0, D'_0, D'_1 \) are derived from the reconstructed \( W'_n \). The notation \( D[P_0] \) refers to the coefficients in \( D \) at the positions specified by the set \( P_0 \), with \( P_0' \) being its complement. The function \( R(D[P_0']) \) randomly selects coefficients from \( D[P_0'] \) such that the number of selected coefficients equals \( |P_0| \). By balancing the number of coefficients in the last two terms of the loss function, we emphasize critical information while regularizing less significant coefficients through random sampling. This approach is also empirically validated in Ablation Section ~\ref{adaptive}.

Our model is initially trained on 10 million samples collected from 19 different datasets (see Section \ref{sec:exp_setup} for details). However, we observed that a substantial portion of this data is skewed towards simple CAD objects, introducing a bias in the training process. This imbalance can cause the model to underperform on more complex or less-represented 3D shapes. To address this issue, we fine-tune the converged VQ-VAE model using an equal number of samples from each of the 19 datasets — a process we call \textit{balanced fine-tuning}. This approach ensures that the model is exposed uniformly to the diverse range of shapes and complexities present across all datasets, thereby reducing the bias introduced by the initial imbalance. Empirically, we find that \textit{balanced fine-tuning} enhances reconstruction results across datasets, as demonstrated in our ablation study (Section~\ref{balance}).

\subsection{Stage 2: Latent Diffusion model}

In the second stage, we train a large-scale generative model with billions of parameters on the latent grid, either as an unconditioned model to capture the data distribution or conditioned on diverse modalities \(  \Theta_n\) (e.g., point clouds, voxels, images). We use a diffusion model within the Denoising Diffusion Probabilistic Models (DDPM) framework ~\citep{ho2020denoising}, modeling the generative process as a Markov chain with two phases.

First, the forward diffusion process gradually adds Gaussian noise to the initial latent code \( Z_n^0 \) over \( T \) steps, resulting in \( Z_n^T \sim \mathcal{N}(0, I) \). Then, the reverse denoising process employs a generator network \( \theta \), conditioned on \( \Theta_n \), to systematically remove the noise and reconstruct \( Z_n^0 \). The generator predicts the original latent code \( Z_n^0 \) from any intermediate noisy latent codes \( Z_n^t \) at time step \( t \), using \( f_\theta(Z_n^t, t, \Theta_n) \approx Z_n^0 \), and is optimized using a mean-squared error loss:
\[
\mathcal{L} = \mathbb{E}_t \left[ \| f_\theta(Z_n^t, t, \Theta_n) - Z_n^0 \|^2 \right]
\]
Here, \( Z_n^t \) is obtained by adding Gaussian noise \( \epsilon \) to \( Z_n^0 \) at time step \( t \) using a cosine noise schedule ~\citep{dhariwal2021diffusion}. The condition \( \Theta_n \) is a latent set of vectors derived from various conditioning modalities, injected into the U-ViT generator ~\citep{hoogeboom2023simple} by using cross-attention and by modulating the normalization parameters in the ResNet and cross-attention layers, as described in ~\citet{esser2024scaling}. This is achieved via condition encoders for different modalities. During training, we apply a small dropout to the condition to implement classifier-free guidance during inference. In the case of unconditional generation, no conditioning is applied. For most input conditions (point clouds, voxels, images, multi-view images, and multi-view depth) we directly train a different conditional generative model for each condition, while for the conditioning on sketch and the single-depth, we take the image-conditioned generative model and fine-tune it with synthetic sketch data and depth data, respectively. For text-to-3D, we fine-tune MVDream~\citep{xu2023dmv3d} to generate six multi-view depth images, as this provides better reconstruction than multi-view images (see experiments in Section ~\ref{image_to_3d}), and then use our model during inference. Further details are provided in the appendix.
\subsection{Inference}
At test time, we begin with a randomly generated noisy latent encoding \( Z_n^T \sim \mathcal{N}(0, I) \) and iteratively denoise it to reconstruct the original latent code \( Z_n^0 \) through the reverse diffusion process, as described in DDPM ~\citep{ho2020denoising}. For conditional generation, we apply classifier-free guidance ~\citep{ho2022classifier} by interpolating between the unconditional and conditional denoising predictions, steering the generation process toward the desired output. This approach allows for greater control over the quality-diversity trade-off. Once the final latent code \( Z_n^0 \) is obtained, we use the pre-trained decoder network of the VQ-VAE from \ref{sec:stage_1_wavelet_vqvae} to generate the final 3D shape in the wavelet form. Subsequently, we apply the inverse wavelet transform to obtain the final 3D shape as an TSDF that can further be converted to a mesh using marching cubes. Notably, we can generate multiple samples for the same conditional input by using different initializations of the noisy latent grid.

\vspace{-0.2cm}
\section{Results}
\vspace{-0.2cm}
\subsection{Experimental Setup}
\label{sec:exp_setup}
\noindent \textbf{Datasets. }
\label{sec:dataset}
Our training data consists of over 10 million 3D shapes, assembled from 19 publicly available datasets, including ModelNet~\cite{vishwanath2009modelnet},
ShapeNet~\cite{chang2015shapenet},
SMPL~\cite{loper2015smpl},
Thingi10K~\cite{zhou2016thingi10k},
SMAL~\cite{zuffi2017smal},
COMA~\cite{ranjan2018coma},
House3D~\cite{wu2018building},
ABC~\cite{koch2019abc},
Fusion 360~\cite{willis2021fusion},
3D-FUTURE~\cite{fu20213d},
BuildingNet~\cite{selvaraju2021buildingnet},
DeformingThings4D~\cite{li20214dcomplete},
FG3D~\cite{liu2021fine},
Toys4K~\cite{stojanov2021using}, 
ABO~\cite{collins2022abo},
Infinigen~\cite{raistrick2023infinite},
Objaverse~\cite{deitke2023objaverse},
and two subsets of ObjaverseXL~\cite{deitke2023objaverse} (Thingiverse and GitHub).
These individual datasets target specific object categories: for instance, CAD models (ABC and Fusion 360), furniture (ShapeNet, 3D-FUTURE, ModelNet, FG3D, ABO), human figures (SMPL and DeformingThings4D), animals (SMAL and Infinigen), plants (Infinigen), faces (COMA), and houses (BuildingNet, House3D). Additionally, Objaverse and ObjaverseXL cover a broader range of generic objects sourced from the internet, covering the aforementioned categories and other diverse objects.
Following \citet{hui2024make}, each of these 19 datasets was split into two parts for data preparation: \(98\%\) of the shapes were allocated for training, and the remaining \(2\%\) for testing. The final training and testing sets were created by merging the corresponding portions from each sub-dataset. Note that we use the entire testing dataset solely for autoencoder reconstruction validation.
We also apply a 90-degree rotation augmentation along each axis, doing the same for the corresponding conditions (point clouds, voxels). 
We also create a balanced training set across these 19 datasets by sampling 10,000 shapes from each. If a dataset contains fewer than 10,000 shapes, we duplicate the data until the target size is reached.

\noindent \textbf{Training Details.}
For optimization and training, we use the Adam optimizer~\cite{kingma2014adam} with a learning rate of $0.0001$ and a gradient clipping value of $1$. For VQ-VAE training, we use a batch size of $256$ with $1024$ codebook embeddings of dimension $4$. We train the network until convergence and then fine-tune the VQ-VAE using a more balanced dataset until it converges again. For the base generative model, we use a batch size of $64$ and train it for $2$ to $4$ million iterations for each modality. Each generative model is trained on a single H100 GPU per condition. We train our model on six conditions: point clouds with $2,500$ points, voxels at $16^3$ resolution, single-view RGB, multi-view RGB with $4$ views, multi-view depth with $4$ views, and multi-view depth with $6$ views. We also fine-tune the single-view model with synthetic sketch data and single-depth data to obtain two more conditions. Additionally, we train an unconditional model beyond these. Finally, we train a large single-view RGB model with 1.4 billion parameters, which we call the WaLa Large model, using 8 H100 GPUs and a batch size of 256. Once this model has converged, we fine-tune it with depth data, using the same number of GPUs and batch size, to obtain the WaLa Depth Large model.

\noindent \textbf{Evaluations Dataset.}
We perform qualitative and quantitative evaluation of our method on Google Scanned Objects (GSO) \citep{downs2022google} and MAS validation data \citep{hui2024make}. 
Importantly, Google Scanned Objects (GSO) is not a part of the dataset detailed in Section \ref{sec:dataset} used to train our model. Consequently, evaluating on Google Scanned Objects (GSO) data assesses the cross-domain generalization of our method. We include all validation objects from the GSO dataset to ensure a broad evaluation.
The MAS validation data is an unseen test set consisting of 50 randomly selected shapes from each of the 19 large-scale datasets mentioned in Section \ref{sec:dataset}.
This ensures that validation data contains all the subcategories like CAD models, human figures, faces, houses, and others, thereby enabling a comprehensive evaluation. We present three metrics for each method on both datasets, the metrics being: (i) Light Field Distance
(LFD) \citep{chen2003visual} which evaluates how alike two 3D models appear when viewed from multiple angles, (ii) Intersection over Union (IoU) ratio, which compares the intersection volume to the total volume of two voxelized 3D objects, and (iii) Chamfer Distance (CD), which measures the similarity between two shapes based on the minimum distance between corresponding points on their surfaces. Note that among these three metrics, for generated shapes that are not aligned (that can occur during generation from conditions such as single images), the most reliable metric is LFD as it is rotation invariant. 

\subsection{Experiments}
We conducted a comprehensive study across various modalities, quantitatively evaluating our method against baselines using four distinct input types: point clouds (Section \ref{sec:pc_to_mesh}), voxels (Section \ref{sec:voxel_to_mesh}), single-view images, and multi-view images (Section \ref{sec:image_to_mesh}). For qualitative analysis, we present the results of all our models, showcasing visual outcomes in Figure ~\ref{fig:teaser} and Figure ~\ref{fig:teaser_2}, and provide additional examples in the appendix as well as on our website: \href{https://autodeskailab.github.io/WaLaProject}{https://autodeskailab.github.io/WaLaProject}. We also report a detailed ablation study in the appendix.

\begin{figure*}[!t]
    \centering
    \includegraphics[width=\textwidth]{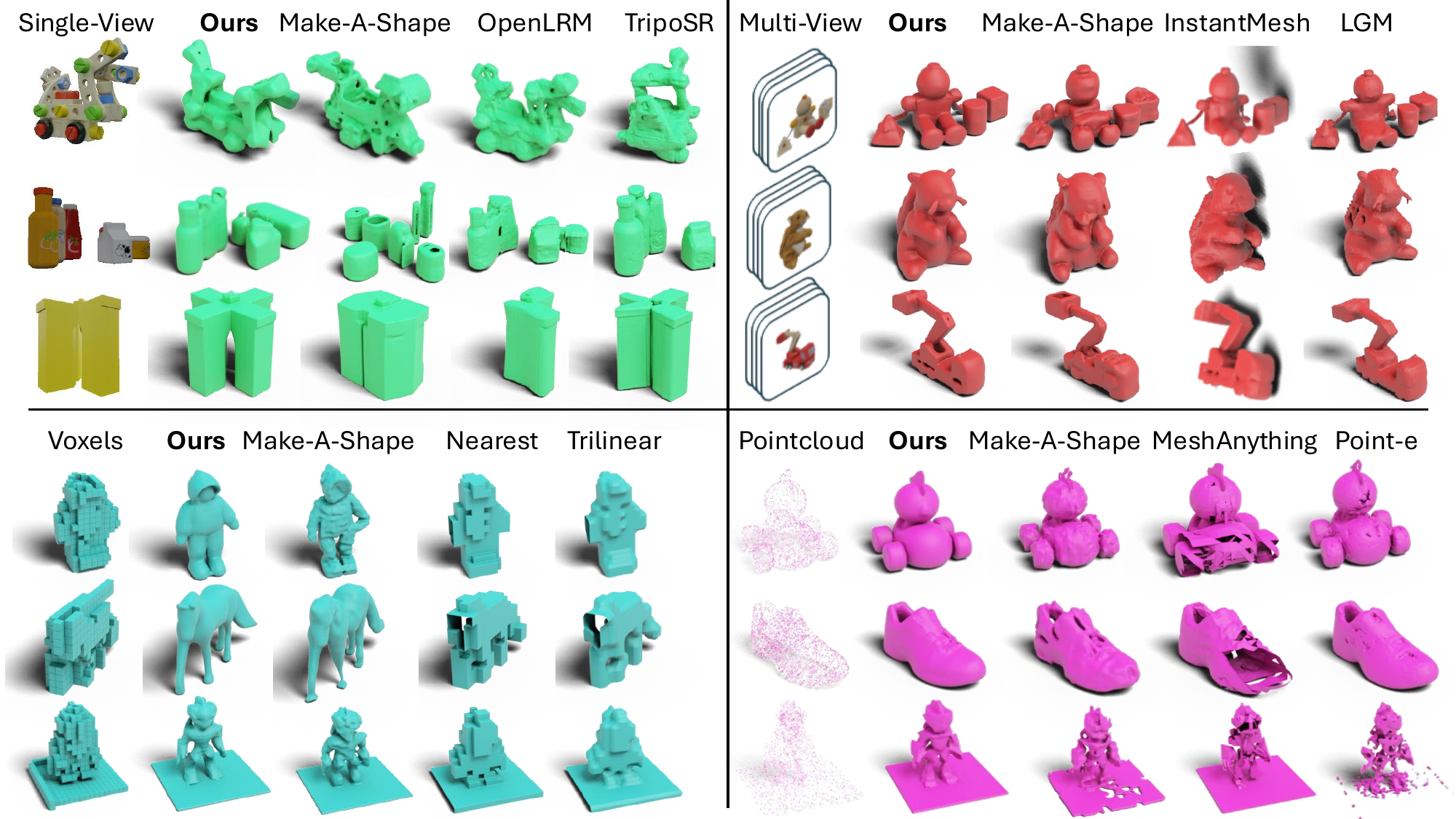}    \caption{Qualitative comparison with other methods for single-view (top-left), multi-view (top-right), voxels (bottom-left), and point cloud (bottom-right) conditional input modalities. \cite{hui2024make, openlrm, TripoSR2024, xu2024instantmesh, tang2024lgm, chen2024meshanything, point-e}
    }
    \label{fig:results}
\end{figure*}

\subsubsection{Point cloud-to-Mesh} \label{sec:pc_to_mesh}
\begin{table}
\centering
{\smaller
\caption{
Quantitative comparison between different methods of point cloud to mesh generation. We present LFD, IOU and CD metrics. Our method, WaLa, outperforms the other methods on both GSO and MAS Validation datasets. } %
\label{tab:pc_to_3d_recon}

\begin{tabular}{c|ccc|ccc}
\toprule
\multirow{2}{*}{\textit{Method}}   & \multicolumn{3}{c}{ GSO Dataset}  & \multicolumn{3}{c}{MAS Dataset}  \\
& LFD $\downarrow$ & IoU $\uparrow$ & CD $\downarrow$ & LFD $\downarrow$ & IoU $\uparrow$ & CD $\downarrow$ \\
\midrule
Poisson surface reconstruction \citep{poisson_recon} & 3306.66 & 0.3838 & 0.0055 & 4565.56 & 0.2258 & 0.0085 \\
Point-E SDF model \citep{point-e} & 2301.96 &  0.6006 &  0.0037 & 4378.51 & 0.4899 & 0.0158 \\
MeshAnything \citep{chen2024meshanything} &   2228.62 & 0.3731 & 0.0064 & 2892.13 & 0.3378 & 0.0091 \\
Make-A-Shape \citep{hui2024make} & 2274.92 & 0.7769 & 0.0019 & 1857.84 & 0.7595 & 0.0036\\
\wala (Ours)  & \textbf{1114.01} & \textbf{0.9389} & \textbf{0.0011} & \textbf{1467.55} & \textbf{0.8625} & \textbf{0.0014} \\
\bottomrule
\end{tabular}
}
\end{table}
In this study, we aim to evaluate the generation of a mesh from an input point cloud containing 2,500 points. We present qualitative results of this task in the bottom right of Figure ~\ref{fig:results} and in rows 1–2 of Figure~\ref{fig:teaser_2}.
To quantitatively assess \wala's performance, we compare it against both traditional and data-driven techniques, as shown in Table~\ref{tab:pc_to_3d_recon}. For the traditional approach, we benchmark against Poisson surface reconstruction, which uses heuristic methods to create smooth meshes from point clouds. For Poisson reconstruction, we need normals, so we estimate them using the five nearest neighbors via O3D~\citep{Zhou2018}. After performing Poisson surface reconstruction, we remove vertices whose density values fall below the 20th percentile to avoid spurious faces.
Additionally, we evaluate our method alongside data-driven generative models such as Point-E~\citep{nichol2022pointe}, MeshAnything~\citep{chen2024meshanything}, and Make-A-Shape~\citep{hui2024make}. For Point-E~\citep{nichol2022pointe}, we utilize its SDF network to estimate the distance field from the point cloud. We also compare our method with MeshAnything~\citep{chen2024meshanything}, a recent transformer-based neural network designed for meshing point clouds. In this case, we use 2,500 input points and follow their hyperparameters and procedure.
Finally, we compare against Make-A-Shape~\citep{hui2024make}, which also generates meshes conditioned on point clouds and has its model open-sourced.

The quantitative results in Table~\ref{tab:pc_to_3d_recon} demonstrate that our method significantly outperforms existing point cloud to mesh generation techniques on both the GSO and MAS validation datasets. These results are despite us not needing normals as in Poisson reconstruction and MeshAnything. Our method can also scale well with data compared to methods like MeshAnything which do not scale well with large face counts. Moreover, our method does not require many surface points to reconstruct a 3D shape, whereas methods like MeshAnything require 8k points (as mentioned in their work) and Point-E requires 4k points.
Qualitatively, our method also outperforms the baselines, as shown in the bottom right of Figure~\ref{fig:results}, and creates smoother shapes with complex geometry, as demonstrated in rows 1–2 of Figure~\ref{fig:teaser_2}.

\subsubsection{Voxel-to-Mesh} \label{sec:voxel_to_mesh}
\begin{table}
\centering
{\small
\caption{
Quantitative evaluation on lower resolution voxel data ($16^3$ resolution) to mesh generation task. Our method, WaLa, surpasses traditional Nearest neighbour and Trilinear upsampling as well as data-centric method like Make-a-Shape.} %
\label{tab:voxel_to_3d_recon}

\begin{tabular}{c|ccc|ccc}
\toprule
\multirow{2}{*}{\textit{Method}}   & \multicolumn{3}{c}{ GSO Dataset}  & \multicolumn{3}{c}{MAS Dataset}  \\
& LFD $\downarrow$ & IoU $\uparrow$ & CD $\downarrow$ & LFD $\downarrow$ & IoU $\uparrow$ & CD $\downarrow$ \\
\midrule
Nearest Neighbour Interpolation & 5158.63 & 0.1773 & 0.0225 & 5401.12 & 0.1724 & 0.0217\\
Trilinear Interpolation   & 4666.85 & 0.1902 & 0.0361 & 4599.97 & 0.1935 & 0.0371 \\
Make-A-Shape \citep{hui2024make} & 1913.69 & 0.7682 & 0.0029 & 2566.22 & 0.6631 &  0.0051 \\
\wala (Ours) & \textbf{1544.67} & \textbf{0.8285} & \textbf{0.0020} &  \textbf{1874.41} & \textbf{0.75739} & \textbf{0.0020} \\
\bottomrule
\end{tabular}
}
\end{table}
In this experiment, we evaluate our proposed method, WaLa, against several baseline approaches for generating 3D shapes from low-resolution voxels with a resolution of $16^3$. Quantitative results are presented in Table~\ref{tab:voxel_to_3d_recon}, while qualitative comparisons are illustrated in the bottom left of Figure~\ref{fig:results} and in rows 3 and 4 of Figure~\ref{fig:teaser_2}.
We evaluate using the GSO and MAS datasets. As detailed in Table~\ref{tab:voxel_to_3d_recon}, \wala~is benchmarked against traditional upsampling techniques (nearest neighbor and trilinear interpolation) and a data-driven approach, Make-A-Shape~\citep{hui2024make}.
For the traditional upsampling baselines, we apply nearest neighbor and trilinear interpolation methods to the $16^3$ voxel grids, followed by the marching cubes algorithm~\citep{lorensen1998marching} to generate the corresponding meshes. In contrast, the data-driven method utilizes the pre-trained voxel-to-mesh model provided by Make-A-Shape.

The results in Table~\ref{tab:voxel_to_3d_recon} demonstrate that \wala~consistently outperforms all baseline methods across various metrics and datasets. Notably, our approach achieves significantly lower LFD and CD values, alongside higher IoU scores, compared to both traditional and data-driven techniques.
These quantitative findings suggest that \wala~not only effectively upsamples 3D shapes to higher resolutions but also produces smoother surfaces and higher-quality meshes by accurately filling in missing details. This holds true even for ambiguous shapes (see Figure~\ref{fig:teaser_2}, third row, columns 3--4) and those with disjoint components (see Figure~\ref{fig:teaser_2}, fourth row, columns 3--4). Furthermore, qualitative assessments in Figure~\ref{fig:results} corroborate our quantitative results, demonstrating that \wala~reconstructs finer geometric features and more precise details than both traditional interpolation methods and existing data-driven approaches.

\subsubsection{Image-to-Mesh} \label{sec:image_to_mesh}
\label{image_to_3d}

\begin{table}
\centering
{\smaller
\caption{
Comparison between different methods on Image-to-3D task (Top) and Multiview-to-3D task (Bottom).
Quantitative evaluation shows that our single-view model excels the baselines, achieving the highest IoU and lowest LFD metrics. Our multi-view model further enhances performance by incorporating additional information. RGB 4, Depth 4, and Depth 6 represents conditioning using RGB images from 4 different views, and depth estimates from 4 and 6 views respectively. Inference time is measured on A100 GPU.
} %
\label{tab:imag_to_3d_recon}
\begin{tabular}{c|c|ccc|ccc}
\toprule
\multirow{2}{*}{\textit{Method}} &Inference  & \multicolumn{3}{c}{GSO Dataset}  & \multicolumn{3}{c}{MAS Val Dataset}  \\
& Time$\downarrow$& LFD $\downarrow$ & IoU $\uparrow$ & CD $\downarrow$ & LFD $\downarrow$ & IoU $\uparrow$ & CD $\downarrow$ \\
\midrule
Point-E \citep{nichol2022point} &$\sim$31 Sec& 5018.73 & 0.1948 & 0.02231 & 6181.97 & 0.2154 & 0.03536\\
Shap-E \citep{jun2023shap} &$\sim$6 Sec& 3824.48 & 0.3488 & 0.01905 & 4858.92 & 0.2656 & 0.02480\\
\multirow{4}{*}{\hspace{-22pt}{\rotatebox[origin=c]{90}{Single-view}}}
One-2-3-45 \citep{liu2023one} &$\sim$45 Sec& 4397.18 & 0.4159 & 0.04422 & 5094.11 & 0.2900 & 0.04036\\

OpenLRM \citep{openlrm} &$\sim$5 Sec& 3198.28 & 0.5748 & 0.01303 & 4348.20 & 0.4091 & 0.01668 \\ 

TripoSR\citep{TripoSR2024} &$\sim$1 Sec& 3750.65 & 0.4524 & 0.01388 & 4551.29 & 0.3521 & 0.03339 \\
InstantMesh\citep{xu2024instantmesh} &$\sim$10 Sec& 3833.20 & 0.4587 & 0.03275 & 5339.98 & 0.2809 & 0.05730 \\
LGM\citep{tang2024lgm} &$\sim$37 Sec& 4391.68 & 0.3488 & 0.05483 & 5701.92 & 0.2368 & 0.07276 \\
Make-A-Shape\citep{hui2024make} &$\sim$2 Sec& 3406.61 & 0.5004 & 0.01748 & 4071.33 &	0.4285 & 0.01851\\ 
\wala~(RGB) &$\sim$2.5 Sec&2509.20 & 0.6154 & 0.02150 & 2920.74 & 0.6056 & 0.01530 \\
\wala~Large (RGB) & $\sim$2.6  Sec & 2473.35 & 0.5984 & 0.02175 & 2562.70 & 0.6610 & \textbf{0.00575} \\
\wala~(depth) & $\sim$2.5 Sec & 2172.52 & 0.6927 & \textbf{0.01301} & 2544.56 & 0.6358 & 0.01213 \\
\wala~Large (depth) &  $\sim$ 2.6 Sec & \textbf{2076.50} & \textbf{0.7043} & 0.01344 & \textbf{2322.75} & \textbf{0.6758} & 0.00756 \\

\hline
\hline

InstantMesh\citep{xu2024instantmesh} &$\sim$1.5 Sec& 3009.19 & 0.5579 & 0.01560 & 4001.09 & 0.4074 & 0.02855 \\
\multirow{4}{*}{\hspace{-30pt}{\rotatebox[origin=c]{90}{Multi-view}}}

LGM\citep{tang2024lgm} &$\sim$35 Sec& 1772.98 & 0.6842 & 0.00783 & 2712.30 & 0.5418 & 0.00867 \\

Make-A-Shape\citep{hui2024make} &$\sim$2 Sec& 1890.85	& 0.7460 & 0.00337 &	2217.25 & 0.6707 & 0.00350 \\

\wala (RGB 4) &$\sim$2.5 Sec& 1260.64 & 0.8500 & 0.00182 & 1540.22 & 0.8175 & 0.00208 \\


\wala (Depth 4) &$\sim$2.5 Sec& 1185.39 & 0.87884 & 0.00164&  1417.40 &0.83313 & 0.00160 \\
\wala (Depth 6) &$\sim$4 Sec&\textbf{1122.61} & \textbf{0.91245} & \textbf{0.00125} & \textbf{1358.82} & \textbf{0.85986} & \textbf{0.00129} \\

\bottomrule
\end{tabular}

}
\end{table}

In this section, we compare \wala~with other state-of-the-art image-to-3D generative models, focusing on both single-view and multi-view scenarios. In the single-view setting, our model generates 3D shapes from a single input image or depth map.
For multi-view generation, we utilize four RGB images or four to six depth images along with their corresponding camera parameters. This approach allows us to evaluate the model's performance under varying conditions, demonstrating the versatility and effectiveness of our generative model in different image-to-3D generation contexts.
Qualitative results for single-view RGB are shown in the top left of Figure~\ref{fig:results} and in rows 5–6 of Figure~\ref{fig:teaser_2}. Conversely, qualitative results for multi-view RGB are displayed in the top right of Figure~\ref{fig:results} and in rows 7–8 of Figure~\ref{fig:teaser_2}. Additional details and results can be found in the appendix and on our website.
Our quantitative results, which assess both quality and inference time on the GSO and MAS validation datasets, are presented in Table~\ref{tab:imag_to_3d_recon}, with the Image-to-3D task results at the top and the multiview-to-3D task at the bottom. We attempted to perform an extensive comparison; however, this proved challenging as many methods are not available as open-source implementations or utilize subsets of the GSO dataset for which the sample lists are not publicly available~\citep{zhang2024clay, siddiqui2024meta, bensadoun2024meta}. Consequently, we chose to use the entire GSO dataset and run open-source models whose code is available for both GSO and MAS.

As demonstrated in Table~\ref{tab:imag_to_3d_recon}, our method consistently outperforms other 3D generation techniques across both tasks.
For the single image-to-3D task, our base RGB model surpasses all baseline methods by a wide margin on most metrics, except for OpenLRM on the CD metric within the GSO dataset. We believe this exception is primarily due to CD's sensitivity to rotation, as many generated shapes may not be perfectly aligned with the ground truth. In contrast, the LFD, which is a rotation-invariant metric, clearly shows significant improvement from WaLa as compared to LRM.
Another noteworthy observation is that the WaLa Large model significantly outperforms the WaLa Base model on the MAS test set but does not show a notable improvement on the GSO dataset. This outcome is expected since we trained on the MAS training set, indicating that increasing the number of parameters may not necessarily enhance generalization to the GSO dataset. Additionally, the single-depth model outperforms the RGB base model, which is intuitive as depth maps provide more comprehensive information about the 3D structure.
Finally, it is important to highlight that our model either outperforms or is comparable to most methods in terms of inference time while delivering significantly better quality. 
A similar trend is observed in the multi-view image-to-3D task, where our model significantly outperforms baseline methods in quality while maintaining similar or better inference times. Another interesting observation is that increasing the number of depth map images improves performance, which again intuitively makes sense as we have more shape information.

Qualitatively, our method generates 3D shapes with complex geometry (see Figure~\ref{fig:teaser_2}, row 8, column 7-8), multiple disjoint components (see Figure~\ref{fig:teaser_2}, row 6, column 3-4), and intrinsic geometric features (see Figure~\ref{fig:teaser_2}, row 5, column 3-4). Additionally, it produces diverse shapes across various object categories, including organic forms (see Figure~\ref{fig:teaser_2}, row 5, column 5-6) and CAD models (see Figure~\ref{fig:teaser_2}, row 7 column 1-2).
Furthermore, our multi-view model outperforms the single-view model, which is intuitively expected due to the additional information provided by multiple perspectives. Our method also visually surpasses other baselines, as demonstrated in Figure~\ref{fig:results}, by capturing more details and creating more complex geometries.
We also apply our multi-view approach to text-to-3D generation, as shown in Figure~\ref{fig:teaser_2}, rows 11–12. For these experiments, we utilize the six-view depth model, which achieves the best reconstruction performance. Additionally, we present visual sketch-to-3D results in Figure~\ref{fig:teaser_2}, rows 9–10. The sketch-to-3D models are obtained by fine-tuning the image-to-mesh model with synthetic sketch data. We also fine-tune the image-to-mesh model with single-view depth data and present visual results in Figure~\ref{fig:teaser_2}
, rows 13–14. Further details about these models are provided in the appendix.

\vspace{-0.3cm}
\section{Conclusion}
\vspace{-0.3cm}

In this work, we introduce Wavelet Latent Diffusion (\wala), a novel approach to 3D generation that tackles the challenges of high-dimensional data representation and computational efficiency. Our method compresses 3D shapes into a wavelet-based latent space, enabling highly efficient compression while preserving intricate details.
WaLa marks a significant leap forward in 3D shape generation, with our billion-parameter model capable of generating high-quality shapes in just 2–4 seconds, outperforming current state-of-the-art methods. Its versatility allows it to handle diverse input modalities, including single and multi-view images, voxels, point clouds, depth maps, sketches, and text descriptions, making it adaptable to a wide range of 3D modeling tasks.
We believe \wala ~sets a new benchmark in 3D generative modeling by combining efficiency, speed, and flexibility. Finally, we release our code and model across multiple modalities to promote further research and support reproducibility within the community.
\vspace{-0.3cm}

\section{Acknowledgement}
\vspace{-0.3cm}

We are deeply grateful to Shyam Sudhakaran and Martynas Pocius for their generous code release, which provided a valuable foundation for this work. We also wish to thank Hilmar Koch for insightful discussions that guided our approach. Special thanks to Justin Matejka and Kendra Wannamaker for their invaluable support in developing the UI interface and advancing our understanding of the model through hands-on experimentation. Dan Ahren’s significant contributions to PR were instrumental, and we are especially thankful for his efforts. Finally, we extend our sincere appreciation to Anthony Ruto, Tonya Custis, Daron Green, and Mike Haley for their steadfast support and encouragement throughout this project. 

\bibliography{references}
\bibliographystyle{iclr2025_conference}

\newpage
\appendix

\section{Additional Results and Details}
For more visual results and detailed information about our model, please visit \url{https://autodeskailab.github.io/WaLaProject}. The code is available at \url{https://github.com/AutodeskAILab/WaLa}.

\section{Architecture details}
In the first stage, we train a convolution-based VQ-VAE using a codebook size of 1024 with a dimension of 4. We downsample the input wavelet tree representation to a $12^3 \times 4$ latent space. Our generative model operates within this latent space by utilizing the U-ViT architecture \cite{hoogeboom2023simple}, incorporating two notable modifications.
Firstly, we do not perform any additional downsampling since our latent space is already quite small. Instead, the model comprises multiple ResNet blocks followed by attention blocks, and then more ResNet blocks at the end, with a skip connection from the initial ResNet block. The attention blocks include both self-attention and cross-attention mechanisms, as described in ~\citep{chen2023pixart}.
Secondly, we modulate the layer normalization parameters in both the ResNet and attention layers, following the approach detailed in ~\citep{esser2024scaling}. This tailored architecture enables our generative model to effectively operate within the compact latent space, enhancing both performance and efficiency.

In this section, we describe the details of the various conditions utilized in our model:

\begin{enumerate}
    \item \textbf{Point Cloud Model}:
    During training, we randomly select 2,500 points from the pre-computed point cloud dataset, which was generated from our large-scale dataset comprising 10 million shapes. These points are encoded into feature vectors using the PointNet encoder \cite{qi2017pointnet}. To aggregate these feature vectors into condition latent vectors, we apply attention pooling as described in \cite{lee2019set}. This process converts the individual points into a latent set vector. Finally, we pass this latent set vector through additional Multi-Layer Perceptron (MLP) layers to obtain the final condition latent vectors.

    \item \textbf{Voxel $16^3$ Model}:
    For voxel-based conditions, we employ a ResNet-based convolutional encoder to process the $16^3$ voxel grid. After applying multiple ResNet layers, the voxel volume is downsampled to reduce its dimensionality to $8^3$. This downsampled volume is then processed with additional ResNet layers, ultimately resulting in the conditional latent vectors. This approach leverages the spatial hierarchy captured by the ResNet architecture to effectively encode volumetric data.

    \item \textbf{Single View Image Model}:
   Our dataset consists of a predetermined set of views for each object. During training, we randomly select one view from this set. The selected view is then processed by the DINO v2 encoder~\citep{oquab2023dinov2} to extract feature representations. The encoder's output serves as the conditional latent vectors, encapsulating the visual information from the single view. It is important to note that we do not train the DINO v2 encoder; instead, we freeze its weights and utilize only the conditional latent vectors.

    \item \textbf{Single View Depth Model}:
    We begin by selecting a checkpoint from a pre-trained Single View Image Model once it has converged and initialize the depth conditioned generative model using the same architecture described in the single-view section. We then fine-tune the model using pre-computed depth data. Throughout this process, we utilize the DINO v2 encoder~\citep{oquab2023dinov2} to obtain the conditional latent vectors while keeping the encoder's weights frozen.

    \item \textbf{Sketch Model}:
    We initialize the model using the architecture described in the single-view section. After the base model converges, we fine-tune it with sketch data. This fine-tuning process involves training the model on sketch representations to adapt the latent vectors, enabling them to capture the abstract and simplified features characteristic of sketches. As in previous cases, the DINO v2 encoder~\citep{oquab2023dinov2} remains frozen. Further details about the sketch data are provided in Appendix~\ref{sketch_data}.

    \item \textbf{Multi-View Image/Depth Model}:
    For multi-view scenarios, we select four viewpoints for the multi-view RGB image model and use configurations with four and six views for the multi-view depth model. These views are carefully chosen from pre-defined angles to ensure comprehensive coverage of the object. Each view is independently processed through the DINO v2 encoder~\citep{oquab2023dinov2}, generating a latent vector for each viewpoint. The latent vectors from all views are then concatenated sequentially, forming a final conditional latent representation structured as a sequence of latent vectors with dimensions corresponding to the number of views and the condition vector size. This approach effectively integrates information from multiple perspectives. It’s also important to note that we keep the DINO v2 encoder frozen in this setup.

    \item \textbf{Text to 3D Model}: In this case, we use the six-view multi-view depth model for 3D generation and the MVDream model for six-view generation from text. The MVDream model is fine-tuned using six-view depth maps, and details are provided in Appendix~\ref{text_3d_details}.
    
    \item \textbf{Uncondtional Model}:
    For the unconditional model, we use the base U-ViT architecture without any conditioning. We only use time to modulate the normalization parameters of the network. Additionally, we do not apply classifier-free guidance.
    
\end{enumerate}

\section{Ablation Studies}
\subsection{VQ-VAE Adaptive Sampling loss analysis} 
\label{adaptive}
In this section, we evaluate the importance of adaptive sampling loss by training two autoencoder models for up to 200,000 iterations: one incorporating the adaptive sampling loss and one without it. The results are presented in the first two rows of Table~\ref{tab:ablation_vq_finetune_model}
. We use Intersection over Union (IoU) and Mean Squared Error (MSE) to measure the average reconstruction quality across all data points. Additionally, we introduce D-IoU and D-MSE metrics, which assess the average reconstruction performance by weighting each dataset equally. This approach ensures that any data imbalance is appropriately addressed during evaluation.

As shown in the table, even after approximately 200,000 iterations, the model utilizing adaptive sampling loss significantly outperforms the one without it. Specifically, the adaptive sampling loss leads to higher IoU and lower MSE values, indicating more accurate and reliable reconstructions.  These results clearly demonstrate the substantial benefits of using adaptive sampling loss in enhancing the performance and robustness of autoencoder models.


\subsection{VQ-VAE Analysis and finetuning Analysis}
\label{balance}
In this section, we examine the benefits of performing \textit{balanced fine-tuning}, as described in the main section of the paper. We conduct an ablation study to determine the optimal amount of finetuning data required per dataset to achieve the best results. The results are presented in the rows following the first two in Table~\ref{tab:ablation_genrative_model}
, utilizing the metrics described above.

Our observations indicate that even a small amount of fine-tuning data improves the IoU and MSE. Specifically, incorporating as few as 2,500 samples per dataset leads to noticeable enhancements in reconstruction accuracy. However, we found that increasing the finetuning data to 10,000 samples per dataset provides optimal performance. At this level, both IOU and Mean Squared Error (MSE) metrics reach their best values, demonstrating the effectiveness of \textit{balanced fine-tuning} in enhancing model performance.

Moreover, the D-IoU and D-MSE metrics confirm that using 10,000 samples per dataset effectively mitigates data imbalance to a certain degree. Based on these findings, all subsequent results in this study are based on using 10,000 finetuning samples per dataset. We believe that an interesting area for future work is to improve data curation to further enhance reconstruction accuracy.






\begin{table}
\centering
{\smaller
\caption{
Ablation study on adaptive sampling as well finetuning of the VQ-VAE model.} %
\label{tab:ablation_vq_finetune_model}

\begin{tabular}{c|c|cccc}
\toprule
\textit{Sampling Loss} & \textit{Amount of finetune data} & IOU $\uparrow$ & MSE $\downarrow$ & D-IOU $\uparrow$ & D-MSE $\downarrow$ \\
\midrule
No\footnotemark[1] & - & 0.91597 & 0.00270 & 0.91597 &  0.00270\\
Yes\footnotemark[1] & - & \textbf{0.92619} & \textbf{0.00136} & \textbf{0.91754} & \textbf{0.00229} \\
\midrule
Yes & - & 0.95479 & 0.00090 & 0.94093 & 0.00169 \\
Yes & 2500 &  0.95966 & 0.00078 & 0.94808 & 0.00149 \\
Yes & 5000 & 0.95873 & 0.00078 & 0.94793 & 0.00149 \\
Yes & 10000 & \textbf{0.95979} & \textbf{0.00078} & \textbf{0.94820} & \textbf{0.00148}\\
Yes & 20000 & 0.95707 & 0.00079 & 0.94659 & 0.00150 \\

\bottomrule
\end{tabular}
}
\end{table}

\footnotetext[1]{Results for the first two rows are based on 200k iterations.}

\subsection{Architecture Analysis of generative model}
In this section, we conduct an extensive study on the architectural design choices of the generative model. Given the high computational cost of training large-scale generative models, we implement early stopping after 400,000 iterations. The results are presented in Table~\ref{tab:ablation_genrative_model}
.
First, we examine the importance of the hidden dimension in the attention layer. It is clearly observed that increasing the dimension enhances performance. A similar trend is noted when additional layers of attention blocks are incorporated. Although the improvement is not pronounced, it is important to mention that these observations are based on only 400,000 iterations.
Finally, we compare the DiT~\citep{peebles2023scalable} architecture to the U-ViT architecture~\citep{hoogeboom2023simple} and find that U-ViT  outperforms DiT. This comparison highlights the superior performance of the U-ViT  architecture in our generative model framework. 

\subsection{Pre-Quant vs post-quant}
\label{pre_or_quant}
In this section, we compare whether it is better to apply the generative model to the grid before or after quantization. We conduct this comparison over 400,000 iterations. The results are shown in Table~\ref{tab:ablation_genrative_model}. These results indicate that pre-quantization performs better.

\begin{table}
\centering
{\smaller
\caption{
Ablation study on the generative model design choices.} %
\label{tab:ablation_genrative_model}

\begin{tabular}{c|c|c|c|ccc}
\toprule
\textit{Architecture} & \textit{hidden dim} & \textit{No. of layers} & \textit{post or pre} & LFD $\downarrow$ & IoU $\uparrow$ & CD $\downarrow$ \\
\midrule
U-VIT & 384 & 32 & pre  & 1523.74 & 0.8211 & 0.001544 \\
U-VIT & 768 & 32 & pre  & 1618.73 & 0.7966 & 0.001540 \\
U-VIT & 1152 & 8 & pre  & 1596.88 & 0.8020 & 0.001561 \\
U-VIT & 1152 & 16 & pre & 1521.81 & \textbf{0.8237} & 0.001573 \\
U-VIT & 1152 & 32 & pre & \textbf{1507.43} & 0.8199 & \textbf{0.001482} \\
DiT & 1152 & 32 & pre   & 1527.16 & 0.8145 & 0.001602 \\
U-VIT & 1152 & 32 & post  & 1576.07 & 0.8176 & 0.001695 \\
\bottomrule
\end{tabular}
}
\end{table}
\section{Sketch data generation}
\label{sketch_data}
\begin{figure}[h]
\centering
\includegraphics[width=\textwidth]{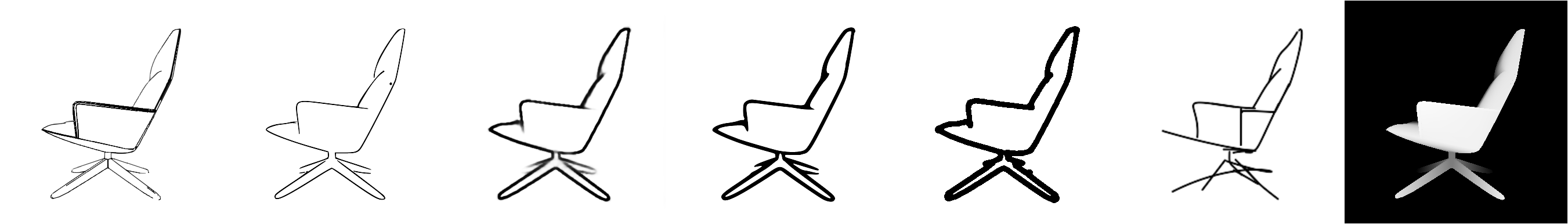}
\caption{The 6 different sketch types. From left to right: Grease Pencil, Canny, HED, HED+potrace, HED+scribble, CLIPaasso, and a depth map for reference. Mesh taken from ~\citep{fu20213d}.}
\end{figure}

We generate sketches using 6 different techniques. In the first technique, we use Blender to perform non-photorealistic rendering of the meshes using a Grease
 Pencil Line Art modifier. The modifier is configured to use a line thickness of
 2 with a crease threshold of 140\textdegree. Since disconnected faces can cause
 spurious lines using this method, we automatically merge vertices by distance using a threshold of 1e-6 before rendering. The second technique takes previously
 generated depth maps and produces sketches using Canny edge detection. We
 apply the Canny edge filter built into \texttt{imagemagick} using a value of 1 for both the blur radius and sigma and a value of 5\% for both the low and high threshold. We then clean the output by running it through the
 \texttt{potrace} program with the flags \texttt{--turdsize=10} and \texttt{--opttolerance=1}. The third technique uses
 HED~\citep{xie2015holistically} in its default configuration, also on depth maps. The fourth technique applies \texttt{potrace} on top of default HED, and the fifth applies HED's \mbox{"scribble"} filter instead. The sixth and
 final technique uses CLIPasso~\citep{vinker2022clipasso} on previously rendered
 color images. We configure CLIPasso to use 16 paths, a width of 0.875, and up to 2,000 iterations, with early stopping if the difference in loss is less than 1e-5.

\begin{figure}[h]
\centering
\includegraphics[width=\textwidth]{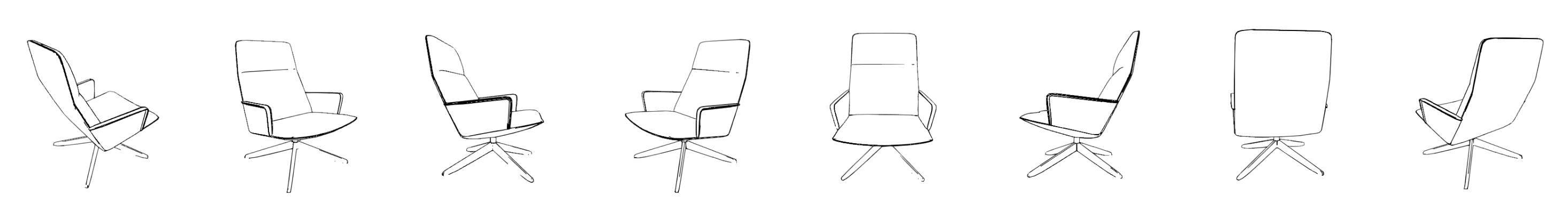}
\caption{The 8 different views for which sketches were generated. Images created using the Grease Pencil technique on a mesh taken from \citet{fu20213d}. The CLIPasso technique was only used on the first, fifth, and sixth views from the left.}
\end{figure}

 For the first 5 techniques, sketches are generated from a total of 8 views: the
 4 views used for multi-view to 3D, plus views from the front, right side, left side,
 and back. For CLIPasso we only generate sketches from the front, right side, and
 left back. Additionally, we only generate sketches from a subset of the 10 million shapes which we constructed by taking up to 10,000 shapes from each of the 20 datasets.

During training we augment the sketches by adding random translation, rotation, and scale in order to improve the model's over-sensitivity to line thickness and padding. We also add random positional noise to the shapes in the SVG drawings
 produced by CLIPasso and \texttt{potrace}. Finally, we add a non-affine cage
 transformation by dividing the image into 9 squares of equal size. We treat the
 four corners of the central square as control points and move each one
 independently, warping the image.
\section{Text-to-3D details}
\label{text_3d_details}
The dataset used for this part contains 330,000 objects, comprising 3D-FUTURE, House3D, Toy4K, ShapeNet-v2, and a filtered subset of Objaverse datasets (filtered by \citep{kant2024spad}). We began by generating captions for this dataset using the Internvl 2.0 model \citep{chen2024internvl}. For each object, we provided the model with four renderings and created two versions of captions by applying two distinct prompts. These initial captions were then augmented using LLaMA 3.1 \citep{dubey2024llama} to enhance their diversity and richness.

Next, we fine-tuned the Stable Diffusion model, initializing it with weights from MVDream \citep{shi2023mvdream}. Utilizing the depth map-text paired data we had collected, we generated six depth maps for each object. To ensure consistency, we identified a uniform cropping box around each object across all depth maps and applied this cropping uniformly to all 6 images. Following the MVDream methodology, we resized the cropped images to 256×256 pixels and employed bfloat16 precision for processing.

During the inference phase, we input text prompts to generate six corresponding depth maps. These depth maps were then used to condition our multi-view depth model, which successfully generated the 3D shape of each object.

\section{Model sizes}
Table~\ref{tab:method_parameters} lists the number of parameters for each of our models.
\begin{table}[ht]
    \centering
    \caption{Number of Parameters for Different Models}
    \label{tab:method_parameters}
    \begin{tabular}{ll}
        \hline
        \textbf{Method} & \textbf{Number of Parameters} \\
        \hline
        Autoencoder Model &  12.9 million \\
        Uncondition Model & 1.1 billion \\
        Single View Model & 956 million \\
        Single View Model Large & 1.4 billion \\
        Depth View Model & 956 million \\
        Depth View Model Large & 1.4 billion \\
        Pointcloud Model & 966.7 million \\
        Multi View Model (Depth and Image) & 956 million \\
        6 view Depth Model & 898 million \\
        Voxel Model & 906.9 million \\
        \hline
    \end{tabular}
\end{table}

\section{Scale and timesteps for different models}
Table~\ref{tab:scale_timestep} lists the classifier-free guidance scales and timesteps used in this paper. These parameters were determined through an extensive grid search on the MAS dataset's validation set. We find that, for most conditions, fewer than 10 timesteps are sufficient, except for the unconditional model. This finding aligns with the results from Make-A-Shape~\citep{hui2024make}, indicating that if the conditioning information is substantial, the diffusion model requires very few timesteps to generate the 3D shape. This is particularly evident in the unconditional setting, where, lacking any shape information hint, the best results are obtained using 1000 timesteps.

\begin{table}[ht]
    \centering
    \caption{Classifier free scale and timestep used in the paper}
    \label{tab:scale_timestep}
    \begin{tabular}{c|c|c}
        \hline
        \textbf{Model} & \textbf{Scale} & \textbf{Timestep}\\
        \hline
        Voxel  & 1.5 & 5 \\
        Pointcloud & 1.3 & 8 \\
        Single-View RGB & 1.8 &   5\\
        Single-View Depth & 1.8 &  5 \\
        Multi-View RGB & 1.3 & 5 \\
        Multi-View Depth & 1.3 & 5 \\
        6 Multi-View Depth & 1.5 &  10 \\
        Unconditional & - & 1000 \\
        \hline
    \end{tabular}
\end{table}
\section{More Visual Results}
\begin{figure*}[!t]
    \centering
    \includegraphics[width=\textwidth]{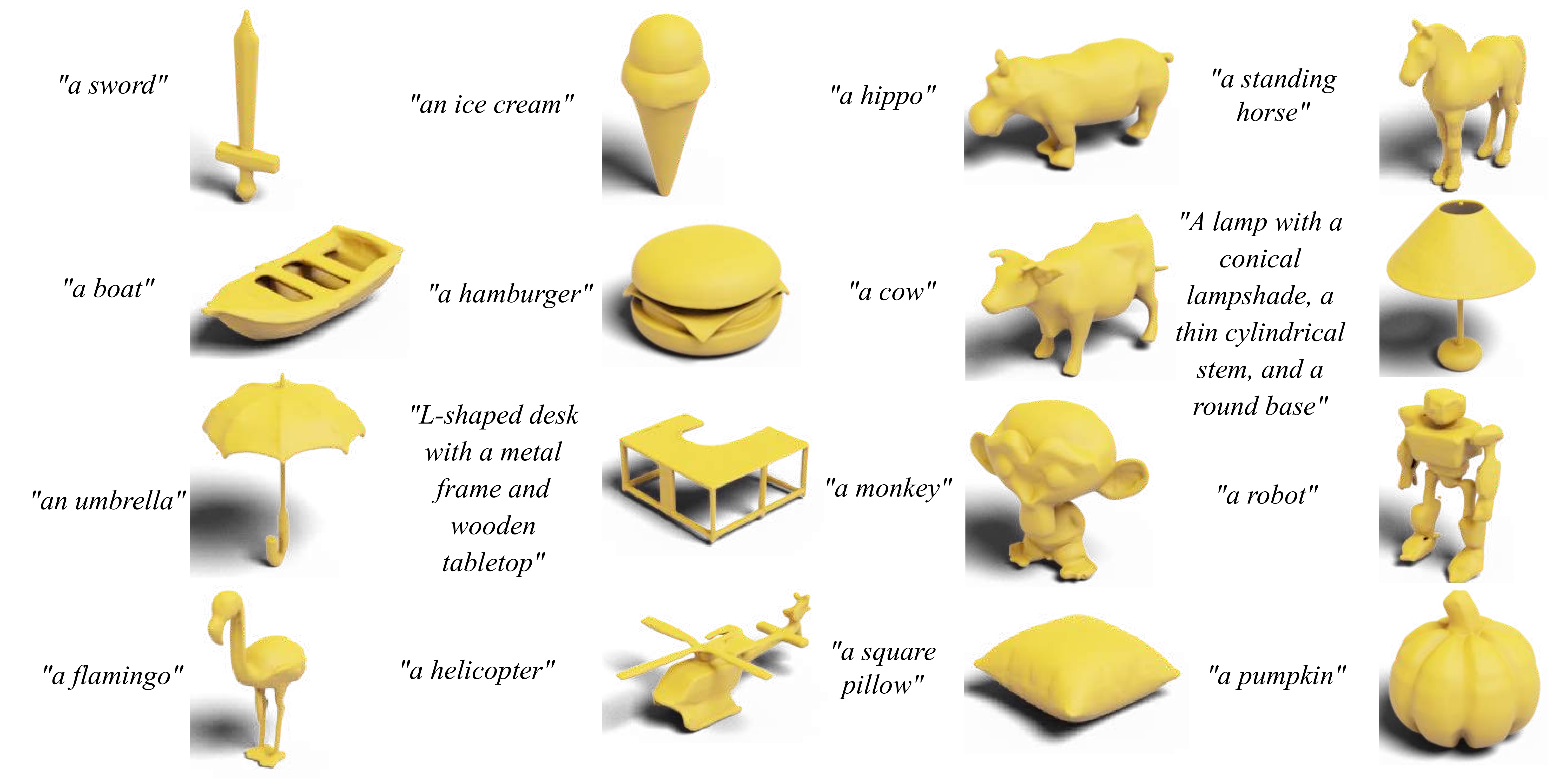}
    \caption{This figure presents more results from the text-to-3D generation task. Each row corresponds to a unique text prompt, with the resulting 3D renderings highlighting the model’s capability to produce detailed and varied shapes from these inputs. 
    }
    \label{fig:text_to_3d}
\end{figure*}

\begin{figure*}[!t]
    \centering
    \vspace{-20pt}
    \includegraphics[width=\textwidth]{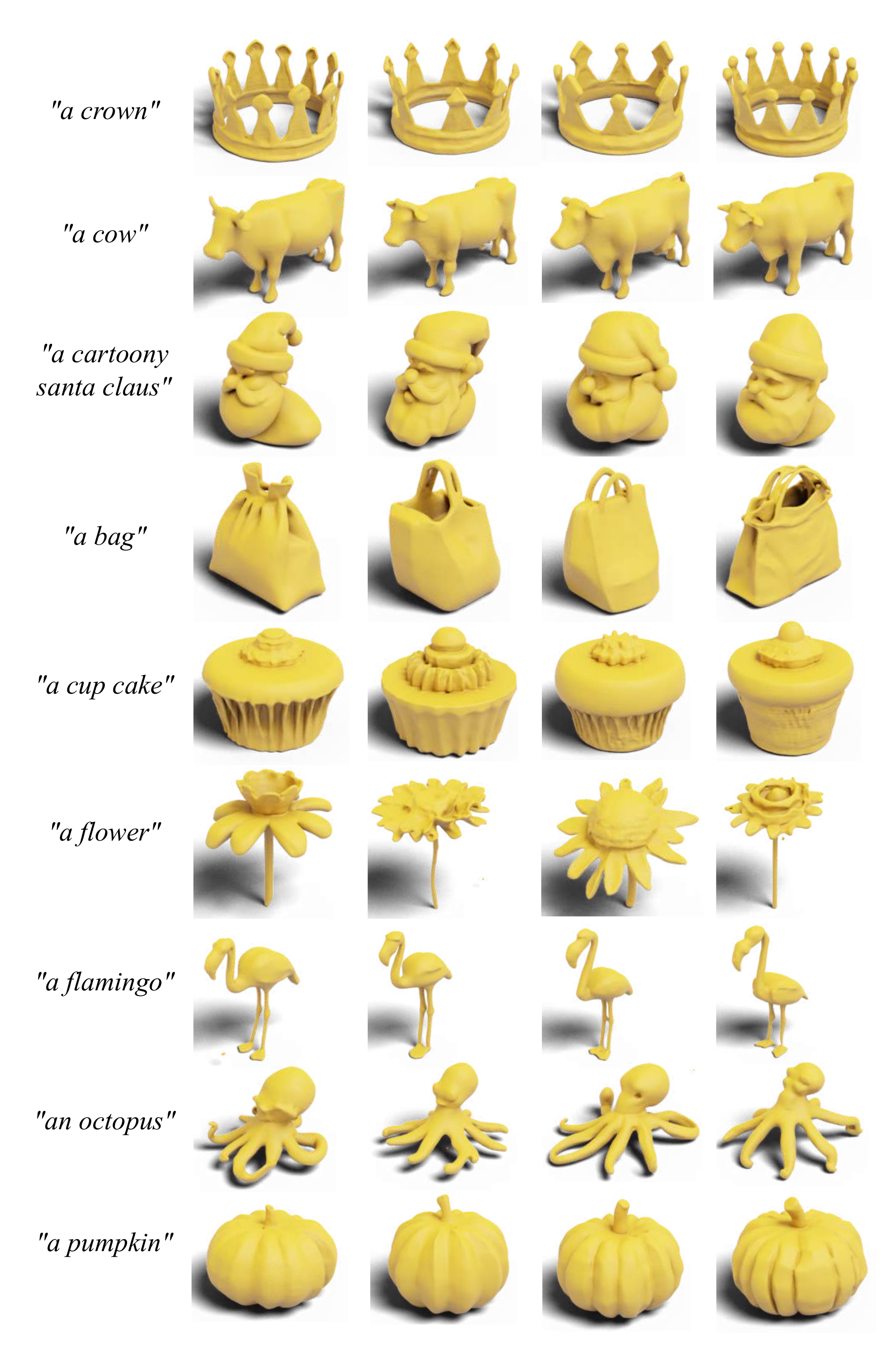}
    \caption{Here, for each caption, four different 3D variations are displayed. This figure emphasizes the model's flexibility in generating multiple distinct outputs for the same text description while maintaining thematic consistency. 
    }
    \label{fig:generation_variety_1}
\end{figure*}
\begin{figure*}[!t]
    \centering
    \includegraphics[width=\textwidth]{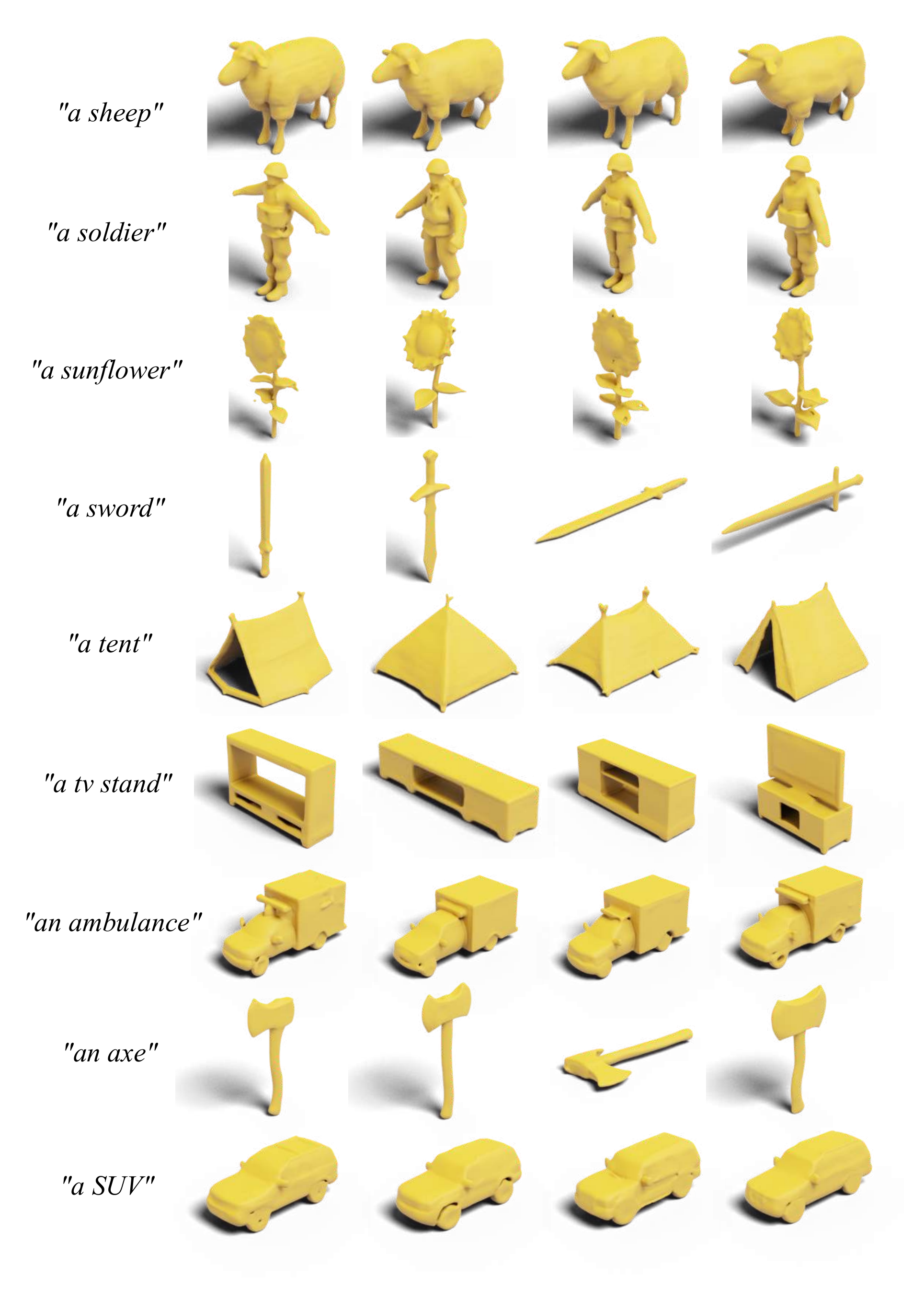}
    \caption{Here, for each caption, four different 3D variations are displayed. This figure emphasizes the model's flexibility in generating multiple distinct outputs for the same text description while maintaining thematic consistency. 
    }
    \label{fig:generation_variety_2}
\end{figure*}

In Figure \ref{fig:text_to_3d}, we present additional text-to-3D generation results, showcasing the diversity and quality of outputs produced by our model. Each result highlights the model’s ability to capture various object details and structures based solely on text prompts. In Figure \ref{fig:generation_variety_1} and Figure \ref{fig:generation_variety_2}, we illustrate the variety in generation for each caption. For each given caption, we display four different generated outputs, demonstrating the model's capacity to create diverse yet semantically consistent results based on the same input description. These figures collectively emphasize the robustness and versatility of our approach in generating 3D content from textual inputs.


\section{Contributions}
Aditya Sanghi: I am the lead author of this paper and contributed significantly to its development. I was responsible for formulating the main research idea, overseeing the dataset generation, and conducting experiments across various modalities. In addition to leading the research team, I coordinated the integration of different sections, ensured cohesion among contributors, and provided guidance throughout the writing process. I also contributed to the drafting and editing of the manuscript and played a key role in creating the figures and visualizations included in the paper. 

Aliasghar Khani: In this paper, I contributed by using vision-language models (VLMs) and large language models (LLMs) to generate captions for over 10 million 3D objects based on their four renderings. I also trained the text-to-multi-view depth generation model utilizing a subset of our dataset. Additionally, I helped create key figures, specifically figures 2, 7, 8, and 9, which visualize and support the paper's findings.

Pradyumna Reddy: Drafting and refining the paper(Introduction, Related Work, Results), along with assistance with visualizations(Fig 4. Single-View and Multi-View), ensuring clear and concise communication. Running multiple baselines (Tab 1. Poisson Surface Reconstruction. Tab 2. Nearest Neighbour Interpolation and Trilinear Interpolation. Tab 3. Single-View, Multi-View InstantMesh and LGM) for quantitative evaluation with current state of the art models. Designing and implementing the project website to enable sharing of a large number of results for each conditioning variable. Active participation in research discussions, focusing on strategies to improve the model's resource efficiency.

Arianna Rampini: Implemented fine-tuning for MVDream, later used in text-to-3D applications. Ran baselines (OpenLRM, TripoSR, Point-E) and computed metrics for quantitative evaluation against state-of-the-art models (Tables 2,3,4). Contributed to presenting the work through paper writing (Image to 3D), proof-reading, and figures creation (Fig 2, 4).

Derek Cheung: Researched and implemented techniques for generating synthetic sketches. Performed fine-tuning experiments using sketches; improved lack of robustness in earlier models by adding additional sketch styles and sketch augmentations. Rewrote data processing scripts to perform sketch generation. Demonstrated that fine-tuning on a balanced subset of the 10m dataset was effective for sketch-to-3d; wrote code for creating and managing balanced dataset subsets. Rewrote distributed inference and evaluation scripts. 

Kamal Rahimi Malekshan: In this project, I contributed by creating the infrastructure for large-scale training and data processing, ensuring smooth workflows. I also prepared the code for release, resolving key issues related to data preprocessing and post-processing. My work focused on enabling efficient model training and large-scale data handling.

Kanika Madan: Contributed towards running of multiple baselines for comparisons with the relevant state of the art methods (Table 1: MeshAnything with different point cloud resolutions, Make-A-Shape; Table 2: Make-A-Shape). Helped with creating the figures (Fig 1, 2, 4, 7, 8 and 9), as well as helped with paper writing and proof-reading.

Hooman Shayani: Co-managed the project alongside Aditya, contributed to writing the paper, and developed Figure 3. Guided the project throughout its development, including idea generation and strategic planning. Collaborated with Derek on sketch generation and provided valuable feedback on various techniques and methodologies. Furthermore, assisted in testing and validating the models and was heavily involved in the code and model releases.



\end{document}